\def\year{2018}\relax
\documentclass[letterpaper]{article} 
\usepackage{aaai18}  
\usepackage{times}  
\usepackage{helvet}  
\usepackage{courier}  
\usepackage{url}  
\usepackage{graphicx}  
\usepackage{comment}
\usepackage{times}
\usepackage{latexsym}
\usepackage[utf8]{inputenc}
\usepackage{amsmath}
\usepackage{amsfonts}
\usepackage{amssymb}
\usepackage{algorithm}
\usepackage{changes}
\usepackage{algpseudocode}
\usepackage{caption}
\usepackage{multirow}
\usepackage{graphicx}
\usepackage{verbatim}
\usepackage{booktabs}
\usepackage{todonotes}
\usepackage[normalem]{ulem}
\useunder{\uline}{\ul}{}

\usepackage{booktabs}
\usepackage{tabularx} 

\usepackage{subfig}
\newcommand{\citet}[1]
{\citeauthor{#1}~\shortcite{#1}}
\newcommand{\citep}{\cite}

\title{Incorporating Discriminator in Sentence Generation: a Gibbs Sampling Method}
\author{Jinyue Su, Jiacheng Xu, Xipeng Qiu\thanks{Corresponding Author}, Xuanjing Huang \\
Shanghai Key Laboratory of Intelligent Information Processing, Fudan University\\
School of Computer Science, Fudan University\\
825 Zhangheng Road, Shanghai, China\\
\{jysu15,jcxu13,xpqiu,xjhuang\}@fudan.edu.cn\\
}

\allowdisplaybreaks

\date{}

\begin{document}

\maketitle

\begin{abstract}

Generating plausible and fluent sentence with desired properties has long been a challenge. Most of the recent works use recurrent neural networks (RNNs) and their variants to predict following words given previous sequence and target label.
In this paper, we propose a novel framework to generate constrained sentences via Gibbs Sampling. The candidate sentences are revised and updated iteratively, with sampled new words replacing old ones. Our experiments show the effectiveness of the proposed method to generate plausible and diverse sentences.
\end{abstract}

\section{Introduction}
Generating meaningful, coherent and plausible text is an increasingly important topic in NLP \cite{Gatt:2017vq}, forming the closed loop of interaction between human and machine. Recent progresses in language modeling, machine translation \cite{Cho:2014iy}, text summarization \cite{Chopra:2016js}, dialogue systems \cite{li2016diversity}, etc., have shown the effectiveness of deep learning methods to understand and generate natural language. Among them, Recurrent Neural Networks (RNNs) and their variants are widely used.  

However, generating natural language remains a challenging task. First, underlying semantic structures in language are complicated and it's hard for sequence models to construct text with rich structure during generation. Second, in many cases, we expect the generated text could satisfy certain properties, such as the polarity of sentiment or the style of writing. Third, annotated corpora for constrained text are limited in terms of scale and diversity. In this paper, we would like to address the challenge of generating short sentences with explicit constraints or desired properties, other than specific tasks like translation or summarization.

Most of the previous works on this task rely on a conditional RNN-based language model as the final step to generate words. The explicit constraint information is encoded into a semantic vector, and a conditional language model decodes a sentence from the semantic vector one word per time step.

On the contrary, we incorporate a discriminator in our method. With the inspiration of revising sentence iteratively, we propose a novel sentence generation framework based on Gibbs Sampling (GS). Our method mainly contains two separate models: a pure \textbf{Language Model}, i.e. doesn't aware of the constraint information, and a \textbf{Discriminator} estimating the probability of a given sentence satisfying the constraints. Then by applying Gibbs Sampling, we can replace current words with new ones iteratively, and the whole sentence is becoming more and more natural. Since the discriminator is guiding this process all the time, the sampled sentences tend to satisfy all the constraints. There is also a \textbf{Candidate Generator} proposing several most likely words at each iteration, thus leave fewer sentences for language model and discriminator to inspect, which results in a speed-up.

Our method holds following advantages compared to conventional methods:
\begin{itemize}
  \item Our method directly accesses to the probability $p(c | w_{1...n})$, i.e. how likely the sentence satisfies all constraints. By checking this probability, our method ensures the significance of every constraint, thus alleviates the problem that some safe but meaningless sentences may be generated. See our experiment results for more details.
  \item Since the language model and the discriminator can be trained separately, we can train the discriminator on labeled data, and train the language model both on labeled and unlabeled data, which enables the framework to work better in a semi-supervised fashion.
\end{itemize}
Also, as a side-product, from the nature of Gibbs Sampling, our method is able to control the length of the generated sentence. And since the two major components are responsible for fluency and correctness respectively, we can easily figure out which part is failing, and then improve it.

To the best of our knowledge, this is the first time Gibbs Sampling is used in generating constrained sentence. 

\section{Background}
In this section, we will introduce two folds of background knowledge: gated recurrent neural networks and 
RNN language model.
\subsection{Gated Recurrent Neural Network}

Gated Recurrent Neural Network(GRU) is a variant of recurrent neural network(RNN), which efficiently implements a dynamic gating mechanism to control the flow of information, aiming to simplify the propagation path of gradients and model the context with memory units. 

A GRU unit takes an input vector $\mathbf{x_t}$ and the previous activation state $\mathbf{h_{t-1}}$ at any time step, and yields an activation state $\mathbf{h_{t}}$ for current time step. $\mathbf{h_{t}}$ is a linear interpolation between previous state $\mathbf{h_{t-1}}$ and the candidate state $\tilde{\mathbf{h_{t}}}$, where the ratio is controlled by a balancing vector $\mathbf{z_{t}}$ called update gate. The model is formulated as follow:
\begin{align}
\mathbf{h_{t}} & = (1-\mathbf{z_{t}}) \odot \mathbf{h_{t-1}} + \mathbf{z_{t}} \odot \tilde{\mathbf{h_{t}}} \\
\mathbf{z_{t}} & = \sigma(W_{z}\mathbf{x_{t}} + U_{z}\mathbf{h_{t-1}}) \\
\tilde{\mathbf{h_{t}}} & = \text{tanh}(W\mathbf{x_{t}}+U(\mathbf{r_{t}} \odot \mathbf{h_{t-1}})) \\
\mathbf{r_{t}} & = \sigma(W_{t} \mathbf{x_{t}} + U_{r} \mathbf{h_{t-1}})
\end{align}
where $\odot$ stands for element-wise multiplication and $\mathbf{r_{t}}$ is the reset gate. The matrices $W$, $U$, $W_{r}$, $W_{z}$, $U_{r}$, $U_{z}$ are trainable parameters. 

\subsection{Recurrent Language Model}


Language models reveal the degree of how much a sequence of words is likely to be a realistic sequence of human language. Formally, let $\Sigma$ be the vocabulary set, and $w_1 , w_2 , ... , w_n$ is a sentence where $w_i \in \Sigma$ is a word in the vocabulary. A language model measures the joint probability by decomposing the words one by one.
\begin{align}
p(w_1...w_n) = \prod_{i=1}^{n} p(w_i|w_{<i}),
\end{align}
where $w_{<i} = (w_1, \cdots , w_{i-1})$.

The prevailing language models are usually based on RNNs~\cite{Mikolov:2010wx,Mikolov:2012bw}. In our model, we adopt GRU~\cite{chung2014empirical} for its efficiency and effectiveness.
The conditional probability $p(w_i|w_{<i})$ can be modeled by GRU,
\begin{align}
p(w_i|w_{<i}) = \phi(\mathbf{x_i} , \mathbf{h}_{i-1}),
\end{align}
where $\mathbf{x_i}$ is the word-vector of the i-th word, and $\mathbf{h}_i$ is the hidden vector at time step $i$. $\phi$ is defined by a single GRU cell.

\section{Task Description}
The task of generating explicitly constrained sentence aims to generate sentence under the constraints of several specific attributes.
More specifically, given $k$ explicit constraints $c=(c_1,\cdots,c_k)$, our goal is to generate a sentence $w=w_1,\cdots,w_n$ which maximize the conditional probability $p(w|c)$
\begin{align}
p(w|c) &= p(w_1,\cdots, w_i|c) \\
&= \prod_{i=1}^n p(w_i|w_{<i},c)
\end{align}

Most of the previous works of generating constrained text are essentially based on a conditional language model. The constraint information is imposed on the network by initial hidden state or concatenated to the input vector at every time step.

The conditional probability $p(w_i|w_{<i},c)$ is computed by
\begin{align}
p(w_i|w_{<i},c) & = \phi(\mathbf{x_i}\oplus \mathbf{a} , \mathbf{h}_{i-1} ),
\end{align}
where $\mathbf{a}$ is the encoding vector of the constraints $c$, and $\oplus$ indicates concatenation operation.

\begin{figure}[t!] 
\centering 
  \includegraphics[width=0.44\textwidth]{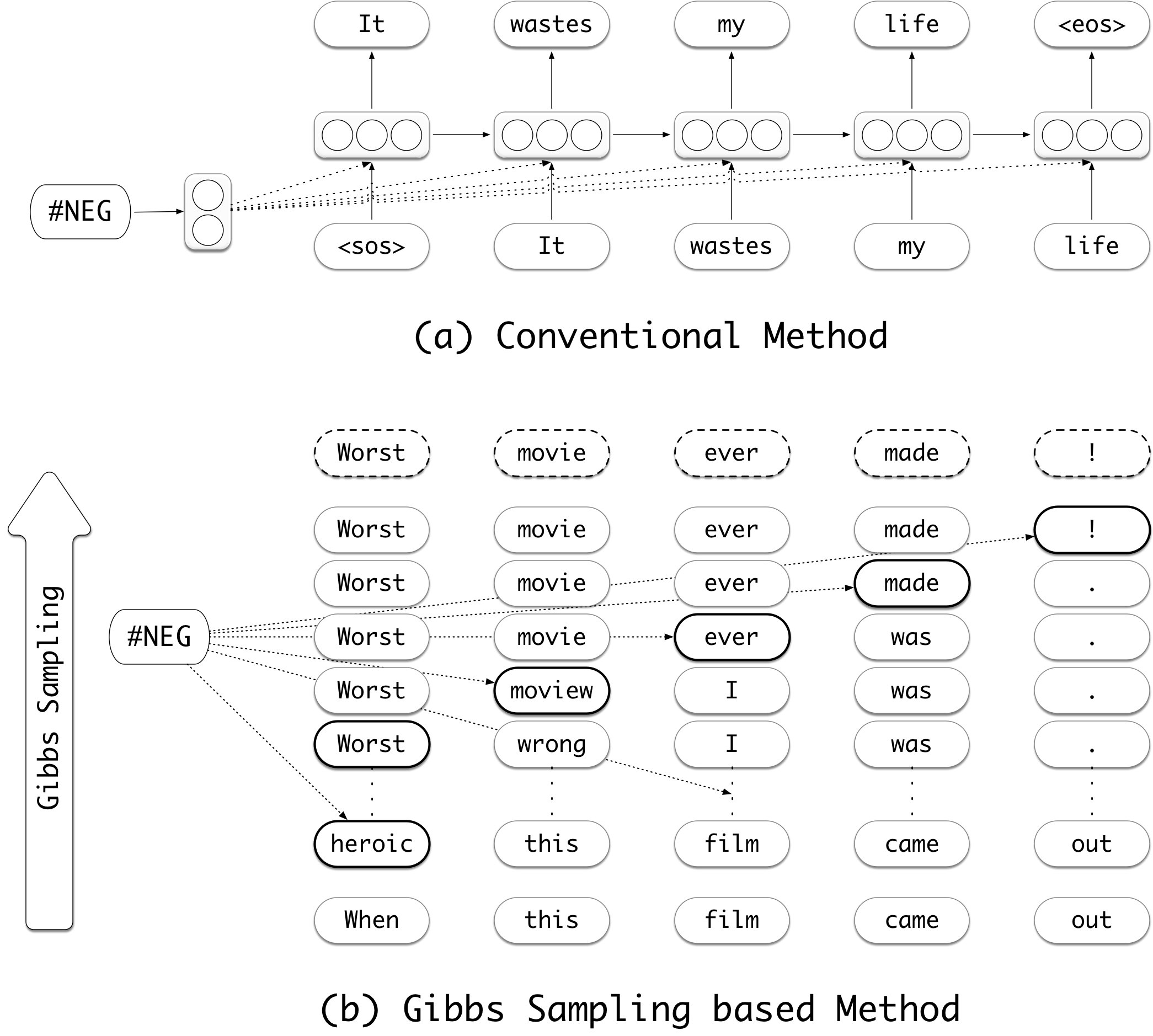}
  
\caption{This figure illustrates how our method works, compared to the conventional method with RNN-based language model.
\textbf{(a) is the diagram of conventional methods}: roll out the language model that is conditioned on the constraints to generate a sentence from left to right. \textbf{(b) is the diagram of our Gibbs Sampling based method}. From bottom to top, ``When this film came out'' is the initial sentence, and ``Worst movie ever made !'' is the final output sentence. Our method replace one word with a new word at a time (words in bolded rounded rectangle). The initial sentence becomes ``heroic this film came out'' after the first modification. In this way, the initial sentence is gradually changed to the final sentence.}
\label{illustrationfig}
\end{figure}

Top part of figure \ref{illustrationfig} gives an illustration of the generating process of conditional RNN.

\section{Generating Constrained Sentence\\ via Gibbs Sampling}

In this section, we introduce our proposed method to generate sentences with imposed constraints.

\subsection{Gibbs Sampling}
Gibbs sampling is a Markov-Chain-Monte-Carlo (MCMC) method, and is widely used in probability inference, such as Latent Dirichlet Allocation (LDA)~\citep{Blei2001LatentDA} etc. 

Gibbs sampling aims to approximately sample a set of random variables from a complicated joint distribution which is intractable to directly sample from. This method starts with the set of variables $x_{1...n}$ randomly initialized. The algorithm goes $T$ turns. In each turn, iterate over all variables $x_i$, and sample new $x_i$ from the full-conditional probability distribution $P(x_i|x_{-i}) \triangleq P(x_i|x_1,x_2...x_{i-1},x_{i+1},x_{i+2},...,x_{n})$. After some turns, the variables $x_{1...n}$ can be seen as sampled from an approximated distribution of the real joint probability distribution.

\begin{algorithm}
\begin{algorithmic}
\Ensure{$x_{1...n}$ randomly initialized}
\For{$t=0$ to $T$}
    \For{$i=1$ to $n$}
        \State sample $x_i$ from $p(x_i|x_{-i})$;
    \EndFor
\EndFor
\State \Return $x_{1...n}$
\end{algorithmic}
\caption{\textbf{Gibbs Sampling}}
\end{algorithm}

\subsection{Our Method}

In this section, we discuss more details of the Gibbs Sampling method in constrained sentence generation.
Our proposed method is based on the MCMC method in \citet{berglund2015bidirectional}, and incorporates the discriminators of constraints to generate constrained sentence. 

When applied to the constrained sentence generation task, the random variables $x_{1...n}$ become the words forming the sentence, i.e. $w_{1...n}$.
\footnote{Note that the number of variables, which is also the length of generated sentence, is invariant during the Gibbs Sampling procedure, which means that we must specify the length at the beginning.}
That is, the algorithm starts from a seed sentence, and repeatedly replaces a word in the sentence with another word sampled from the full-conditional distribution. Figure \ref{illustrationfig} gives an illustration of this procedure.

More formally, let $c_1,c_2,...c_k$ be the constrains, and $c$ represents all the constraints.

In constrained sentence generation task, we want to sample sentences from distribution $p(w_{1...n} | c)$. Thus the full-conditional distribution used in Gibbs Sampling is $p(w_i | w_{-i} , c) \triangleq p(w_i | w_{<i} , w_{>i}, c)$. Use Bayes theorem, we get

\begin{align}
& p(w_i | w_{-i} , c) \\
&= p(w_i | w_{<i} , w_{>i}, c) \\
&= \frac{p(w_{1...n} , c)}{p(w_{<i}, w_{>i},c)}\\
& \propto p(w_{1...n} , c) \\
& = p(w_{<i}, w_{>i}) \cdot p(w_i | w_{<i}, w_{>i}) \cdot p(c | w_{1...n}) \\
& \propto  p(w_{i} | w_{<i} , w_{>i}) \cdot p(c | w_{1...n}). \label{eq:model}
\end{align} 

Eq (\ref{eq:model}) can be decomposed into two parts dealt with language model and discriminator respectively.

\paragraph{Language Model}
The first part is $p(w_{i} | w_{<i} , w_{>i})$.
And there are two ways to deal with it. 

In a direct fashion, we can simply train a Bidirectional Recurrent Neural Network here to estimate the probability $p(w_{i} | w_{<i} , w_{>i})$. More concretely, a forward RNN takes the prefix as the input and a backward RNN takes the suffix. Let the final states be $\mathbf{h}_f$ and $\mathbf{h}_b$ respectively. Then the probability is calculated in the following way

\begin{align}
p(w_{i} | w_{<i} , w_{>i}) = \mathbf{v}_{w_{i}}^T (\mathbf{W} [\mathbf{h}_f ; \mathbf{h}_b] + \mathbf{b}),
\end{align}
where the vector $\mathbf{v}_{w_{i}}$ is the one-hot vector of the word $w_{i}$, and $\mathbf{W}$, $\mathbf{b}$ are trainable parameters.

Or, we can apply Bayes theorem again, and get 
\begin{align}
& p(w_i | w_{<i} , w_{>i}) \\
& = \frac{p(w_{1...n})}{P(w_{<i}, w_{>i})} \\
& \propto p(w_{1...n}) \\
& = \prod_{t=1}^{n} P(w_t | w_{<t}).
\end{align}

Then the probability takes the form $p(w_t | w_{<t})$ which is exactly the same with language model.

In our experiments we use the latter way to estimate the probability $p(w_{i} | w_{<i} , w_{>i})$, since the latter model can estimate the probability of the sentence $p(w_{1...n})$ as well, which is needed to decide which sentence to output.

\paragraph{Discriminator}
The second part of Eq (\ref{eq:model}) is the conditional probability of constraints $c$ given $w_{1...n}$.
For example, if the constraint is the sentiment of movie review, then this term is the probability of the sentence $w$ holding positive / negative sentiment. We can train a classifier to discriminate the sentiment and estimate the probability.

In our experiments, we use GRU-based discriminators. We build a forward RNN to take the sentence as input, and a backward RNN to take the reverse of the sentence. Let $h_{f}$ and $h_{b}$ be the final states of the two RNNs respectively. Then pass the two vector to a two-layer MLP, and get the probability of the sentence over all categories of the constraint.


When there are multiple constraints, we can make the independent assumption that any two constraints are independent to each other, which is almost correct in many situations. Then we have $p(c | w_{1...n}) = \prod_{i=1}^{k} p(c_{i} | w_{1...n})$. Thus we can train $k$ discriminators independently on the $k$ constraints.


We do the transduction above because the task to model a complicated distribution has been decomposed to model two much simpler distributions. We can make use of language model to estimate the first term on the right hand side in equation(\ref{eq:model}), and discriminators to estimate the second term.

By doing this, we can sample many sentences from the distribution $p(w_{1...n} | c)$, and choose one as the output. In our experiment, we choose the sentence where $\forall i \in [1,k] , p(c_{i} | w_{1...n}) > threshold$ with the largest $p(w_{1...n})$.

\subsection{The Candidate Generator}

Still, there remains two major problems in the Gibbs Sampling based method.

One is that the probability $p(c | w_{1...n})$ sometimes is not well-defined on every single choice of $w_{1...n}$, since there are many meaningless ``sentences'' in the $|\mathcal{V}|^{n}$ space, and it is hard to say a meaningless sequence of words has positive or negative sentiment.

The second problem is the time efficiency issue. In most applications, the size of vocabulary $|\mathcal{V}|$ is around 10k. It's impossible to calculate the probability $p(w_i | w_{<i} , w_{>i}) \cdot p(c | w_{1...n})$ for every possible value of $w_{i}$, especially the second term.

To alleviate these two problems, we propose the candidate generator. Basically, the candidate generator takes the constraints and the prefix along with the suffix as inputs, and output $\mathbf{k}$ words as most probable candidates for the $i$-th position. By replacing the original $i$-th word with these words, we get $\mathbf{k}$ candidate sentences. Then, the language model and the discriminator will go through these sentences, calculate corresponding probability, and sample one from them.

In this way, every sentence sent to the discriminator will be more natural, and the number of probability calculation will be reduced to $\mathbf{k}$ in every iteration of inner loop.

This method is still very time consuming after the speed-up, though. The discriminator needs to look all these $\mathbf{k}$ sentence-with-one-word-replaced over, and estimates the probability.

In our experiments, for simplicity, the candidate generator calculates the probability distribution $p(w_{i} | w_{<i} , w_{>i})$ using a Bidirectional Recurrent Neural Network, and outputs $\mathbf{k}$ words with the highest probability.

\begin{table}[t]
\centering

\begin{tabular}{ll}

\toprule
Hyper parameters               & Value \\
\midrule
Hidden-units              & 200   \\
Word-vec-size             & 200   \\
Constraint-embedding-size & 10    \\
Turns in Gibbs Sampling   & 100   \\
Fixed sentence length     & 8    \\
Burn-in turns               & 10 \\
Threshold                 & 0.6  \\
Candidate word number($\mathbf{k}$)    & 5 \\
Sentences sampled in RS   & 800 \\
Beam size                 & 300 \\

\bottomrule
\end{tabular}
\caption{Hyper-parameters setting}\label{fig:hyper}
\end{table}

\begin{table}[t]
\centering
\begin{tabular}{lll}
\toprule
Model       & SST-2             & Product          \\ 
\midrule
Pure LM       & 72.83 & 88.01  \\

Conditional LM     & 66.64 & 109.33 \\

Candidate Generator & 46.02  & 64.64\\

\bottomrule

\end{tabular}
\caption{The perplexity on test datasets.}
\label{testperplexity}
\end{table}

\begin{table}[t]
\centering\small
\begin{tabular}{lll}
\toprule
Model             & SST-2       & Product \\ 
\midrule
Sentiment discriminator & 0.738 & 0.784 \\ 
Domain discriminator    & /           &     0.866   \\
\bottomrule
\end{tabular}
\caption{Classification accuracy of discriminators.}
\label{classificationaccuracy}
\end{table}

\begin{table*}[t]
\small
\centering
\begin{tabular}{l|l|l|l|p{0.2\textwidth}}
Sentiment                 & \textbf{Gibbs Sampling}         & \textbf{Beam Search}        & \textbf{Reject Sampling} & \textbf{\citet{hu2017controllable}}     \\ \hline
\multirow{3}{*}{negative} & Yes , the movie is too bad .    & It ’s a bad movie .             & It 's a bad movie . & the acting was also kind of
hit or miss .         \\ \cline{2-5} 
                          & One of the worst of the year .  & It ’s a bad film .              & An awful film . & the movie is very close to
the show in plot and characters             \\ \cline{2-5} 
                          & The film is a little too long . & It ’s not very funny .          & It 's not a good movie .  & i wo n’t watch the movie   \\ \hline
\multirow{3}{*}{positive} & This is a movie that is funny . & A very funny movie .            & A fascinating documentary . & his acting was impeccable  \\ \cline{2-5} 
                          & It ’s a great kind of movie .   & It ’s a very funny movie .      & It 's a good movie . & this is one of the better
dance films        \\ \cline{2-5} 
                          & One of the year ’s best films . & One of the year ’s best films . & It 's a great film . & i hope he ’ll make more
movies in the future

\end{tabular}
\caption{The generated sentences for movie review. We can observe that Beam Search method tends to output sentences that are safe but less meaningful: the domain tag seems to be ignored. }
\label{resultssst2}
\end{table*}

\begin{table*}[t]
\centering\small
\begin{tabular}{c|p{0.3\textwidth}|p{0.25\textwidth}|p{0.2\textwidth}}
\hline
{Sentiment} &
\textbf{Gibbs Sampling} & \textbf{Beam Search}  & \textbf{Reject Sampling}
\\ \hline
\multicolumn{4}{c}{Books}                                                                                                                                                                      \\ \hline
{negative}       & {this book is a complete waste of time}             & {i was disappointed}     & this is the worst book ever           \\ \hline
{positive}       & {this book is a great book to read}                 & {highly recommended}     & i love this book                      \\ \hline
\multicolumn{4}{c}{DVDs}                                                                                                                                                                       \\ \hline
{negative}       & {after watching this movie i was very disappointed} & {i was disappointed}     & i was disappointed                    \\ \hline
{positive}       & {this is the best movie i’ve ever seen}             & {highly recommended}     & i love this movie                     \\ \hline
\multicolumn{4}{c}{Electronics}                                                                                                                                                               \\ \hline
{negative}       & {and i would not recommend it to anyone}            & {don't waste your money} & it was just fine                      \\ \hline
{positive}       & {if you are looking for a good camera}              & {highly recommended}     & it works great                        \\ \hline
\multicolumn{4}{c}{Kitchen Appliances}                                                                                                                                                         \\ \hline
{negative}       & {but i will never buy this product again}           & {don't waste your money} & it does not work well                 \\ \hline
{positive}       & {so i bought this one for my husband}               & {i love it}              & i would highly recommend it to anyone \\ \hline
\end{tabular}
\caption{The generated sentences for product reviews.}
\label{resultsproduct}
\end{table*}

\section{Experiments}

In this section, we investigate the performance of the proposed method on two review datasets: a movie review dataset (SST-2)~\citep{socher2013recursive}
 and a product review dataset~\citep{blitzer2007biographies}.

The task is to generate constrained sentence. On movie review, the sentiment is designated, and on product review, both the domain and sentiment are constrained.

\subsection{Dataset}

SST-2 is a dataset which contains movie reviews with two classes of sentiment label : ``negative'' or ``positive''. SST-2 is from the Stanford Sentiment Treebank\footnote{http://nlp.stanford.edu/sentiment}. 

The Product dataset\footnote{https://www.cs.jhu.edu/˜mdredze/datasets/sentiment/} contains reviews from Amazon in four domains: Books, DVDs, Electronics and Kitchen appliances. Each review has a sentiment label of ``positive'' or ``negative''. We split documents in this dataset to sentences for training language model. The sentiment discriminator and domain discriminator is still trained on document level.

\subsection{Baseline}

We use the conventional \textbf{Beam search (BS)} method and a \textbf{Reject sampling (RS)} method for generating constrained sentence as baselines.

\textbf{Beam Search} is a widely used approximate algorithm to find the sentence with maximum likelihood estimated by the language model. This algorithm keeps a set of generated prefixes. The size of the kept set is called beam size $m$. At each time step, these prefixes would be extended by one word, and the top $m$ new prefixes would be kept for the next iteration.

\textbf{Reject Sampling} method works as follows: randomly sample sentences from language model with no constrain information for $\hat{T}$ times, and output the sentence where $\forall i \in [1,k] , p(c_{i}|w_{1...n}) > threshold$ with the largest $p(w_{1...n})$. This method randomly pick words within top-10 words with most probability at each time step, in order to improve the performance.
This method shares the same discriminator and language model with the Gibbs Sampling method.

\subsection{Settings}

All Recurrent Neural Networks used in our experiments are GRU with $200$ hidden units. The sentence was fed to the network by replacing words with embeddings. 
And the inputs to the conditional language model are concatenated to extra constraint-embedding vectors at each time step. Words occurs fewer than 10 times in dataset are replaced with UNK. After the preprocessing, the vocabulary size of SST-2 is about 3700 and Product about 7000.

The beam size of Beam Search is set to 300 in order to make sure all the methods take about the same amount of time to generate a sentence. The number of random sampled sentences in the reject sampling method is set to 800 for the same reason.

$T$ is the turns executed in Gibbs Sampling, and $n$ is the fixed sentence length. The seed sentence is modified from a segment from training data consisting of $8$ words: if $n > 8$, use UNK for padding; if $n < 8$, only use the prefix. And the length of burn-in period is set to $10$ turns. Burn-in is a term used in Gibbs Sampling context, and it means that we don't care the samples in first several turns.

As a reminder, Pure LM is the language model used in Gibbs Sampling and Reject Sampling, and Conditional LM is the conditional language model used in Beam Search.

Table \ref{fig:hyper} lists the setting of hyper-parameters. Test error of all components are shown in Table \ref{testperplexity} and \ref{classificationaccuracy}. It is weird that the test perplexity of conditional language model on Product dataset is higher than pure language model while on SST-2 the thing is opposite. This may due to the fact that there are more kinds of constraints on Product ($2 \times 4$ where 2 for sentiment and 4 for domain) than on SST-2 (2 for sentiment).

\begin{table}[t]\small
\centering
\begin{tabular}{l}
\toprule
It 's too bad .                                   \\
The movie is funny .                              \\ \midrule
I ca n't recommend it .                           \\
It 's a great movie .                             \\ \midrule
The film is just too bad .                        \\
It 's a very good film .                          \\ \midrule
One of the worst of the year .                    \\
This is one of the best films .                   \\ \midrule
Yes , the film is a bad movie .                   \\
It 's a great piece of a movie .                  \\ \midrule
This is a movie that is n't very funny .          \\
This is a good movie in its own way .             \\ \midrule
It 's very funny , but it 's a mess .             \\
It 's a good movie with a lot of humor .          \\ \midrule
The film is one of the worst movies of the year . \\
It 's not very funny , but it 's worth seeing .   \\
\bottomrule

\end{tabular}
\caption{The generated sentences with length from 5 to 12 for movie reviews.}
\label{resultslength}
\end{table}

\subsection{Results}
Some illustrations of the generated sentences are shown in Table \ref{resultssst2} and \ref{resultsproduct}. Since \citet{hu2017controllable} have done experiments on SST dataset, we also present some of their results here. We also change the fixed length of sentence, and the generated results are shown in Table \ref{resultslength}.

To make a quantitative evaluation, we use BLEU~\citep{papineni2002bleu}
to estimate the quality of the generated sentences.

The BLEU score is a method to automatically evaluate the quality of sentence generation, which has been shown to be highly correlated to human judgment when applied to machine translation. The BLEU score calculates the precision of n-gram matching between generated sentences and references. A brevity penalty term is also used to penalize too short generated sentences. 

We use the BLEU-4 (the geometric mean from 1 gram to 4 grams) in experiments.
More concretely, the score of a single generated sentence is the average BLEU score matching this sentence to all sentences with the same label from the dataset (including train, dev, and test split). 80 randomly generated sentences are used to get the average score of a method. The number of sentences in each category is equal.

The Beam Search method and Reject Sampling inherently tend to generate short sentence. This is unequal in the comparison, since the Gibbs Sampling will definitely output sentences of length 8, which is longer than most of the sentences generated by the two baselines. To take the length out of consideration, we also calculate BLEU score without brevity penalty.

Results are shown in Table \ref{bleusst2} and \ref{bleuproduct}. ``RANDOM'' stands for the average BLEU score of 80 randomly chosen sentences from the same dataset with the designated labels.

\begin{table}[t]
\centering
\begin{tabular}{lll}
\toprule
method                  & w/ penalty  & w/o penalty    \\ 
\midrule
RANDOM                  & \textbf{4.338} & 5.888 \\ \hline
GS          & 2.060 & \textbf{8.302}  \\
BS             & 1.290 & 6.059 \\
RS & 0.384 & 4.029 \\
\bottomrule
\end{tabular}
\caption{BLEU scores ($\times 10^{-3}$) on SST-2 dataset. }
\label{bleusst2}
\end{table}

\begin{table}[t]
\centering
\begin{tabular}{lll}
\toprule
method                  & w/ penalty  & w/o penalty    \\ 
\midrule
RANDOM                  & \textbf{2.093} & 2.854 \\ \hline
GS          & 1.476 & \textbf{4.750}  \\
BS             & 0.218 & 4.556 \\
RS & 0.369 & 3.323 \\
\bottomrule
\end{tabular}
\caption{BLEU score (Product) $\times 10^{-3}$}
\label{bleuproduct}
\end{table}

\subsection{Discriminative Efficiency}

We further check the ``discriminative efficiency'' of our method. Here the ``discriminative efficiency'' means the ratio of sentences satisfying the constraints among all sampled sentences. In other words, the probability estimated by the discriminator satisfy: $\forall i \in [1,k] , p(c_{i} | w_{1...n}) > threshold$. Since the output is the sentence with the most likelihood among all valid sentences, the quality of output is highly related to the ratio of valid sentence samples.

\begin{table}[t]
\centering
\begin{tabular}{lll}
\toprule
ratio                   & SST-2       & Product     \\ 
\midrule
RANDOM                    & 0.5         & 0.125       \\ \hline
GS          & \textbf{0.846} & \textbf{0.682} \\ 
RS & 0.478 & 0.114  \\
\bottomrule
\end{tabular}
\caption{Valid Sentence Ratio}
\label{validtable}
\end{table}

In Table \ref{validtable}, some statistical data of valid sentence ratio is presented. The data is collected when the two methods generate the 80 sentences in the last subsection. There are about 57k(after burn-in) and 64k sentences sampled respectively.

RAND is the probability of a randomly sampled sentence from dataset to have the exact label. As shown in the table, the sampling process is skewed to the label that was designated.

\begin{figure}[t] 
\centering\includegraphics[width=0.45\textwidth]{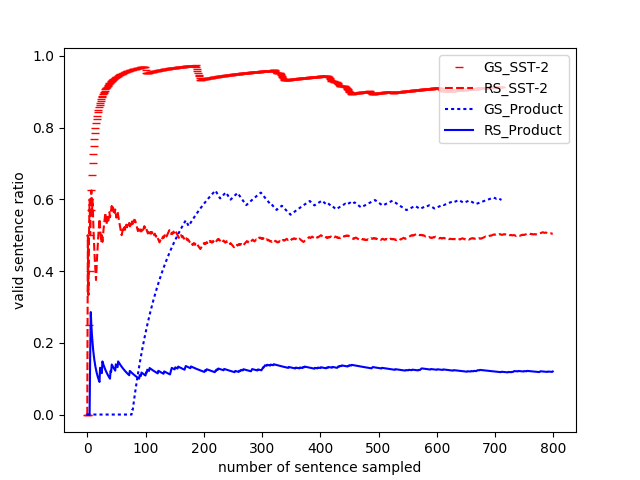} 
\caption{Valid Sentence Ratio}
\label{validratio}
\end{figure} 

Figure \ref{validratio} shows how the ratio goes during generation procedure. The curves presented here are sampled with the constraint [positive] on movie review and [electronics , positive] on product review.

\subsection{Log-likelihood per Word}

\begin{figure}[t] 
\centering\includegraphics[width=0.45\textwidth]{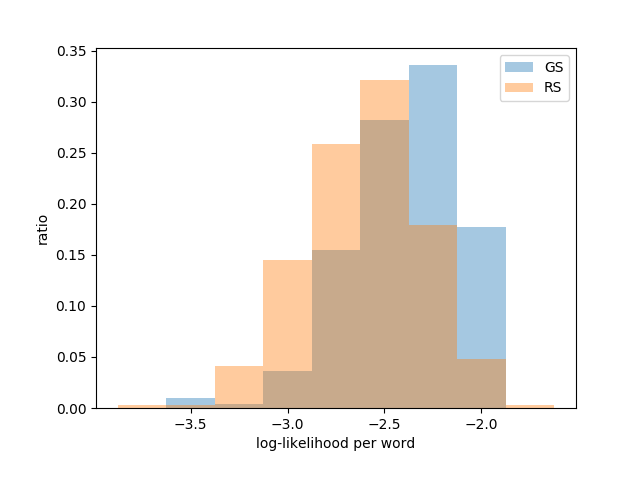} 
\caption{Log-Likelihood per word on movie review. Better sentences lie on the right part of the x-axes.}
\label{likelihoodsst2}
\end{figure}

\begin{figure}[t] 
\centering\includegraphics[width=0.45\textwidth]{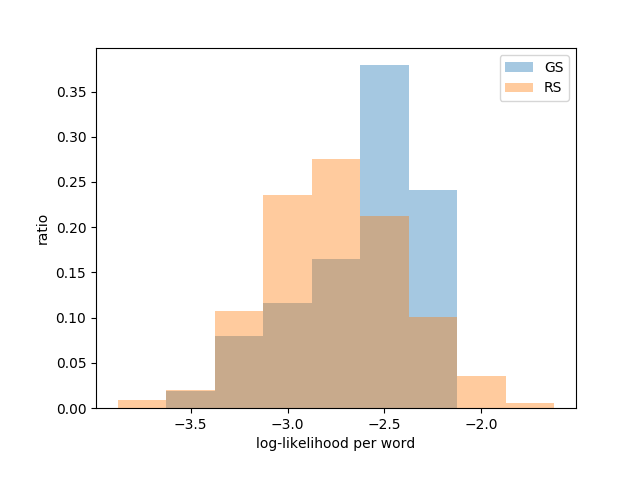} 
\caption{Log-Likelihood per word on product review. Better sentences lie on the right part of the x-axes.}
\label{likelihoodproduct}
\end{figure}

We also checked the log-likelihood per word of sentences sampled by the Gibbs Sampling method, and the Reject Sampling method. As shown in Figure \ref{likelihoodsst2} and \ref{likelihoodproduct}, we can see that the sentences generated by Gibbs Sampling are considered to be more realistic by the pure language model, which is the same in both methods.

The shown figures \ref{likelihoodsst2} and \ref{likelihoodproduct} are sampled with the constraint [positive] on movie review and [electronics , negative] on product review.

\subsection{Failure Cases}

In our experiments, we observe several kinds of failure. Some of them are listed below.

\begin{itemize}
\item Broken sentence segments. For example, \emph{this book is not written in this book} seems to be an unnatural concatenation of two sentence segments. This is due to the failure of language model.
\item Lack of context. For example, \emph{but i am very pleased with this book} is not a complete sentence. The word \emph{but} indicates a turning of sentiment, but there is no context. This is also a kind of language model failure.
\item Failure of discriminator. \emph{i would think this is a good product} is generated with label [Electronics, negative].
\end{itemize}

\subsection{Analysis}

Our experiments show that the proposed Gibbs Sampling method achieves a significantly better BLEU score with or without brevity penalty, and can generate sentences with controllable length. 

It's interesting that Gibbs Sampling achieves a higher BLEU score than RANDOM when there is no brevity penalty, which shows the diversity in the dataset.
We also empirically show that Gibbs Sampling is more ``discriminative efficient'' than conventional sample method, since the valid sentence ratio is marginally higher.

And the log-likelihood per word of sampled sentences are higher in Gibbs Sampling than conventional sampling. It seems that the language model given contexts are more precise, and candidate generator works just well.

\section{Related Work}

Recently, benefiting from the powerful neural language model, neural network based text generation achieved great successes on many NLP tasks, including machine translation \cite{Cho:2014iy,bahdanau2014neural,luong2015effective}, summarization \cite{Chopra:2016js,Miao:2016ul}, and image caption \cite{vinyals2015show,xu2015show}.
These tasks can be formalized as generating text with the conditional language model. The condition is implicitly modeled by distributed representation of several constraints. The constraints are different in different tasks. For example, the constraint in machine translation is the semantics of the source language, while the constraint in image caption is the image content. 

Another line of related work focuses on generating sentence from explicit attributes, such as product reviews. 

\citet{mou2015backward} use two directional RNNs to generate sentence containing specific word.
\citet{donglearning} use network architecture similar to that in NMT: an attribute encoder outputs the constraint embedding, and the sequence decoder with attention to softly align the attributes and generated words. 
\citet{hu2017controllable} use the variational autoencoder (VAE) \cite{Kingma:2013tz}, a regularized version of standard autoencoder, and employs the wake-sleep algorithm to train the whole model in a semi-supervised fashion. They are also using the dataset SST, though the experiment settings are very different from ours(the constraints in their works are tense and sentiment). Some of sentences generated by their method are shown in \ref{resultssst2} for comparison. \cite{Goyal:2017wi} build a language model capable of generating emotionally colored conversational text. These methods feed the constraint information as extra features to the networks, while our method use discriminators to get control of the sentence.

And recently, \citet{guu2017generating} propose a method that generates sentences by prototype-then-edit. The pipeline is very similar to ours, though technically very different.

\section{Conclusion and Future Works}

In this paper, we propose a novel Gibbs-Sampling-based method to generate constrained sentence. The proposed method doesn't directly model the distribution $p(w_{1...n} | c)$. Instead, the method makes use of several separately trained models, and apply Gibbs Sampling to generate sentence. Experimental results show that this method can yield diverse and meaningful sentences.

In the future, we plan to further investigate the relationship between the seed sentence segment and the final output sentence. We have observed that they may share some sentence pattern, but the procedure of transferring the seed sentence to the output sentence still remains mysterious. Besides, we plan to apply the proposed framework to some sequence-to-sequence task, chatbot for example. Lastly, we are wondering whether it will be an improvement, to replace a segment in the sentence at a time, in order to gain more fluent intermediate sentences. Also, this framework can work in a semi-supervised fashion, which we haven't thoroughly examined. We also left this for future work.

\section*{Acknowledgement}
We would like to thank the anonymous reviewers for their valuable comments. The research work is supported by the National Key Research and Development Program of China (No. 2017YFB1002104), Shanghai Municipal Science and Technology Commission (No. 17JC1404100), and National Natural Science Foundation of China (No. 61672162).

\bibliography{emnlp2017,jcxu}
\bibliographystyle{aaai}

\end{document}